\documentclass{egpubl}

 \JournalSubmission    
 \electronicVersion 

\ifpdf \usepackage[pdftex]{graphicx} \pdfcompresslevel=9
\else \usepackage[dvips]{graphicx} \fi

\PrintedOrElectronic

\usepackage{t1enc,dfadobe}

\usepackage{egweblnk}
\usepackage{cite}
\usepackage{amsmath}

\title[]%
      {Lightweight Markerless Monocular Face Capture with 3D Spatial Priors}

\author[]{Shridhar Ravikumar}

\teaser{
\includegraphics[width=\textwidth]{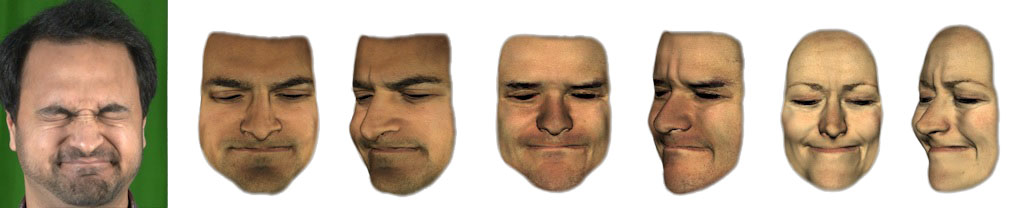}
           \includegraphics[width=\textwidth]{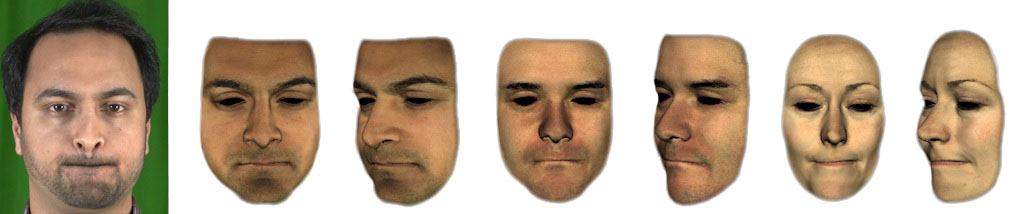}   
 \centering
  \caption{Results from our solver re-targeted onto multiple face models. As seen in the side views, our method is able to handle ambiguities in depth, inherent in monocular input, and achieve results that are physically plausible even in areas with occlusions like around the lips.}
\label{fig:banner}
}

\begin{document}
\maketitle

\begin{abstract}
We present a simple lightweight markerless facial performance capture framework using just a monocular video input that combines Active Appearance Models for feature tracking and prior constraints on 3D shapes into an integrated objective function. 2D monocular inputs inherently lack information along the depth axis and can lead to physically implausible solutions. In order to address this loss of information, we enforce a constraint on our objective function within a probabilistic framework that uses preexisting animations obtained from accurate 3D tracking systems, thus achieving more plausible results. Our system fits a Blendshape model to tracked 2D features while also handling noise in estimation of features and camera parameters. We learn separate constraints for the upper and lower regions of the face thus maintaining flexibility. We show that using this approach, we can obtain significant improvement in tracking especially along the depth dimension. Our method uses easily obtainable prior animation data. We show that our method can generate convincing animations using only a monocular video input. We quantitatively evaluate our results comparing it with an approach using a monocular input without our spatial constraints and show that our results are closer to the ground-truth geometry. Finally, we also evaluate the effect that the choice of the Blendshape set has on the results of the solver by solving for a different set of Blendshapes and quantitatively comparing it with our previous results and to the ground truth. We show that while the choice of Blendshapes does make a difference, the use of our spatial constraints generates results that are closer to the ground truth.
\end{abstract}

\section{Introduction}

\noindent Over the last few years many methods for facial performance capture have been developed. These methods range in complexity from active marker based approaches with multiple cameras, head-mounted devices and controlled environments \cite{Williams_1990_PerformanceDrivenFacialAnimation,Vicon,Fyffe_2011_ComprehensiveFacialPerformanceCapture,Furukawa_2009_Dense3DMotionCapture,Fyffe_2014_DrivingHighResolutionFacialScans,Bradley_2010_HighResolutionPassivePerformanceCapture,Beeler_2011_AnchorFrames,IanMatthews_2011_InteractiveRegionBased,Raitt_2004_Gollum,Seol_2012_SpaceTimeExpressionCloning,Guenter_1998_MakingFaces,KiranBhat_2013_ContoursBasedCapture,Valgaerts_2012_LightweightBinocular,Alexander_2008_DeformationDrivenPolynomialDisplacementMaps} to passive markerless monocular approaches \cite{Kemelmacher_2014_TotalMovingFace,TimCootes_2001_AAM,Saragih_2011_RealtimeAvatarAnimationFromSingleImage,ChenCao_2013_3DShapeRegression,ChenCao_2014_DisplacedDynamicExpression,ChenCao_2015_RealtimeHighFidelity,Garrido_2013_MonocularCapture}. Then there are methods that use depth sensing devices like the Microsoft Kinect and others \cite{Zhang_2008_SpaceTimeFaces,Weise_2009_FaceOff,Weise_2011_RealtimePerformanceBased,HaoLi_2013_OnTheFlyCorrectives,Bouaziz_2013_OnlineModelingForRealtimeAnimation}. Although many of these methods provide good quality animations, the time taken to setup the environment including the time taken to apply markers on the actor's face can be a hindrance to the quick application of these methods, similarly the capture devices used may be custom built and not easily available or cheap enough for general consumer use. That being said, there is ample data that exists from these legacy systems, freely available online, captured from multiple people performing a range of facial movements and speech.
One of the problems with 2D monocular inputs is that information along the depth axis is inherently missing and this proves to be a challenge. Even though the solver may minimize the error in the objective function, the true 3D shape may be ambiguous. In this work, we present a lightweight approach that achieves good quality solves using just a monocular input while exploiting this existing data from legacy systems and using it to learn a prior in order to make sure physically implausible results aren't generated. 
Our contributions are as below:

\begin{itemize}
\item We present a lightweight monocular markerless capture method that achieves good quality animation parameters and does not require special equipment or complex training phases.
\item We exploit easily available prior animation data obtained from 3D tracking systems and use this in a density estimation framework to regularize our objective function and generate more plausible results. We enable further flexibility by learning separate prior constraints for the upper and lower face regions.
\item We combine initial estimates of 2D landmark points on the face based on an ensemble of regression trees, with an Active Appearance Model for improved accuracy.
\item We handle noise in input 2D features and in the estimation of camera extrinsic parameters thus eliminating jitter in the resulting animation.
\item Finally we evaluate the effect that the choice of Blendshapes has on the results of the solver in general and show that while it affects the results, incorporating the spatial priors generates results that are closer to the ground truth.
\end{itemize}

\section{Related Work}

\noindent Many different approaches for facial performance capture have been demonstrated in the last decade or so and can broadly be categorized into different types mainly based on the representation that is used for the underlying animation model and also based on the input devices used for the capture process.

\noindent Mesh propagation approaches deform a single high quality mesh of the actor's face through the sequence in order to generate the animation based on motion capture data.  Bradley et al.~\cite{Bradley_2010_HighResolutionPassivePerformanceCapture} propagate a mesh by applying optical flow over multiple cameras and combine both 2D video and 3D point clouds in order to obtain very high resolution captures. Fyffe et al. \cite{Fyffe_2011_ComprehensiveFacialPerformanceCapture,Fyffe_2014_DrivingHighResolutionFacialScans} use five high speed cameras combined with gradient illumination patterns  in order to get extremely high resolution captures. They also make use of multiple static scans in order to account for high resolution detail. Beeler et al. \cite{Beeler_2011_AnchorFrames} use anchor frames in order to reduce drift in optical flow. \cite{Kemelmacher_2014_TotalMovingFace} automatically reconstruct the 3D face shape of the subject from an unconstrained data-set of images obtained over the internet and track the face through a monocular video sequence by finding the optimal scene flow parameters. They then add in fine scale details using a shape from shading approach. Other mesh propagation approaches include \cite{Guenter_1998_MakingFaces,Borshukov_2005_MatrixReloaded,Valgaerts_2012_LightweightBinocular}.

\noindent On the other hand, parametric model based approaches optimize for the values of the parameters through the sequence in order to best capture the movement of the face. Statistical models, such as those based on Principle Component Analysis \cite{TimCootes_2001_AAM,BlanzVetter_1999_3DMorphableModels,IanMatthews_2011_InteractiveRegionBased} provide an orthogonal set of basis which can then be combined linearly in order to generate the animation. Blendshape models consist of a set of facial expressions as a basis and these expressions are combined in a linear fashion in order to generate different expressions. This is arguably the most popular approach for representing animation owing to its intuitiveness and relative ease of use. \cite{JPLewis_2014_PracticeAndTheoryOfBlendshapes} gives an excellent in-depth analysis of Blendshapes. \cite{Weise_2009_FaceOff,Bouaziz_2013_OnlineModelingForRealtimeAnimation,Garrido_2013_MonocularCapture,HaoLi_2010_ExampleBasedFacialRigging,HaoLi_2013_OnTheFlyCorrectives,ChenCao_2013_3DShapeRegression,ChenCao_2014_DisplacedDynamicExpression,Deng_2006_CrossMappingMotionCapture} all make use of Blendshape models combined with either multi-view, monocular or depth sensing input devices.
\noindent \cite{Thies_2015_RealtimeFacialReenactment} presented a method for real-time facial reenactment using an RGB-D device. They use a statistical model for the identity, expression and albedo of the face and optimize for the best model parameters in an analysis by synthesis approach. They model the scene lighting using spherical harmonic basis by assuming that the light source is distant and the scene is predominantly lambertian. They optimize for the parameters in real-time using a data parallel GPU solver.

\noindent In this work, our objective is to capture the movement of the actor's face using a lightweight method that doesn't require expensive equipment or lighting apparatus and using only a single monocular off-the-shelf camera while simultaneously addressing the issue of missing depth information. In light of this, we will mainly discuss previous work that is similar to our approach in these regards.

\noindent Our approach is most similar in spirit to Garrido et al. \cite{Garrido_2013_MonocularCapture}. They present a lightweight approach that makes use of a single monocular video input and are able to generate a high quality output animation with fine scale details using a Blendshape model. They track a few sparse landmark feature points on the face reliably using forward and backward optical flow combined with automatic key-frame selection based on local binary patterns for robustness. The pose and facial expressions are estimated from these sparsely tracked points on the face in an iterative fashion. Temporally coherent dense motion fields tracked from video combined with a smoothness constraint is then used in order to refine the pose and facial expression. Fine scale details are then added on top using a shape from shading approach. Although their method produces good results, as it is a monocular system, it doesn't account for inherent loss of information along the depth dimension. Our method uses a prior constraint in order to regularize the data which alleviates the error due to the lack of depth information. Also our method doesn't depend on optical flow across the sequence, but only on the current and previous frames and thus can be implemented in an online fashion.

\noindent Cao et al. \cite{ChenCao_2013_3DShapeRegression} presented a method for obtaining real-time performance capture from monocular input.
Similar to our approach, they track 2D points but instead of fitting directly to the 2D features, they train a user specific two-level boosted regressor trained on labelled 2D points and corresponding 3D shapes, in order to map from 2D to 3D features at run-time. They then fit a Blendshape model to the obtained 3D features by iteratively solving for transformation parameters and expression weights. Their method requires a training phase where images of the user in different poses and expressions are captured. \cite{ChenCao_2014_DisplacedDynamicExpression} extend the previous work to be independent of a user and instead learn a regressor from public image databases. They infer both the 2D facial landmarks and the 3D shape of the face simultaneously. Their algorithm adapts to the user's face at run-time by solving for the user-specific Blendshapes and the expression co-efficients in an iterative manner. \cite{ChenCao_2015_RealtimeHighFidelity} enhance a low-resolution tracked mesh with medium-scale wrinkle details which are generated by local regressors trained on high-resolution scan data. They require a one-time training phase in order to learn the mapping from UV space to vertex offsets and can be applied to an unseen actor at runtime. While their method obtains very impressive and detailed results, the training process is quite involved and requires generation of multiple guess-truth pairs for the parameters for each training image in order to train the DDE regressor as explained in \cite{ChenCao_2014_DisplacedDynamicExpression}.

\section{System Overview}
\label{sec:SystemOverview}

\noindent Our pipeline fits a Blendshape model to automatically tracked 2D feature points in the video. In order to generate our Blendshapes (Sec.\ref{sec:AutomaticBlendshapeGeneration}), we first deform the neutral expression mesh from an existing template Blendshape model to a 3D scanned face of the actor to obtain the mesh of the actor in a desired topology. We then automatically generate the user-specific Blendshape expressions for this actor. Our 2D landmark features are initially obtained using the method of \cite{Kazemi_2014_OneMilliSecond} giving us 68 distinct landmarks (Sec.\ref{sec:2DFeatureDetection}). These landmarks are detected per frame and although they provide a good starting point, the detections are not accurate enough for our purposes and give unacceptable results especially for extreme facial expressions. In order to address this, we use the Fast Simultaneous Inverse Compositional algorithm of \cite{Tzimiropoulos_2013_FastAAM}, trained on a few images of the user which is applied on top, using the results from \cite{Kazemi_2014_OneMilliSecond} as an initialization. This gives us more robust landmark detections especially in extreme facial expressions. We solve for the camera extrinsic parameters using the Perspective-n-Point approach with the Levenberg-Marquardt algorithm for non-linear optimization (Sec.\ref{sec:CameraParameters}). In order to address the inherent noise from estimating landmarks and camera parameters per frame, we smooth the noise in the 2D features and the camera parameters using a Kalman filter (Sec.\ref{sec:KalmanFilter}). In order to regularize our solve results, we make use of a prior energy constraint (Sec.\ref{sec:PriorConstraints} and Sec.\ref{sec:DataBinning}) that estimates the probability of the solution and factors it into the objective function (Sec.\ref{sec:ObjectiveFunctionWithConstraints}) resulting in more plausible shapes. Our optimization function then solves for the optimal coefficients of the Blendshapes that minimize the re-projection error of the landmark vertices whilst taking into consideration its likelihood. In Section \ref{sec:Quant} we quantitatively compare our results to the ground-truth and show that our method generates more accurate results. Finally in Section \ref{sec:EmilyComparison}, we evaluate a completely different set of Blendshapes in order to analyze the effect that the choice of Blendshapes has on the solver result, and also quantitatively show that our method still generates more accurate results.

\section{2D Facial Feature Detection}
\label{sec:2DFeatureDetection}
\noindent To obtain our initial 2D features, we first use the algorithm of \cite{Kazemi_2014_OneMilliSecond} implemented within the Dlib library \cite{Dlib},  to detect 68 landmarks on the face on a frame by frame basis. The algorithm uses an ensemble of regression trees to accurately predict the optimal displacement of each landmark at each level, based on differences in pixel intensity around that landmark. We trained the ensemble of regression trees using the images from the HELEN, LFPW, AFW, IBUG and 300 faces In-the-wild databases \cite{300FacesInWildDatabase_2016,300Faces_2013_SemiautomaticMethod,300Faces_FirstLandmarkLocalizationChallenge_2013}, to give a total of 4213 training images with a wide variety of pose, lighting and shape variations. We used a cascade depth of 10, tree depth of 10, 500 trees per cascade, a feature pool of 400 and 50 test splits (see \cite{Kazemi_2014_OneMilliSecond} for details). This gives us reasonable initial detections of landmarks on the user's face but isn't accurate or robust enough for the application of performance capture as shown in the accompanying video. The landmark detections are incorrect during extreme expressions and this leads to unacceptable solve results.

\noindent In order to improve upon this, we further train an Active Appearance Model \cite{TimCootes_2001_AAM} using the Fast Simultaneous Inverse Compositional approach of \cite{Tzimiropoulos_2013_FastAAM}, on a few select images of the user performing a few facial expressions. We used 15 facial expressions that included a few that elicit the extreme range of the user's facial movements (jaw-open, lip-swing) and also a few challenging expressions that generate lip and eye occlusions (pucker, squinch). We then use the result of the previous step as a starting point and then use the AAM to improve landmark tracking through the sequence. In our experiments, this gives us much more robust detections and also covers the full range of the user's expressions as shown in the accompanying video. The results of this landmark detection are further fed into the Kalman filter to account for discrepancies between frames and to remove noise in an online fashion.

\begin{figure}[t]
    \centering          
    		\includegraphics[width=1.5in]{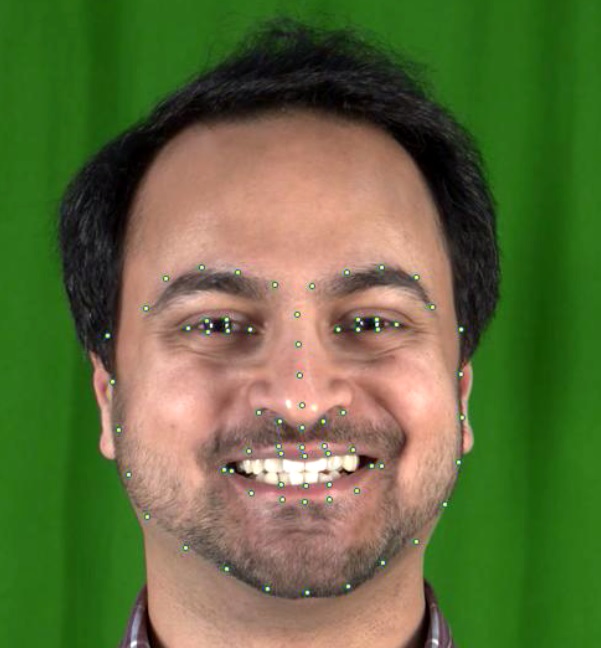}  
            \includegraphics[width=1.38in]{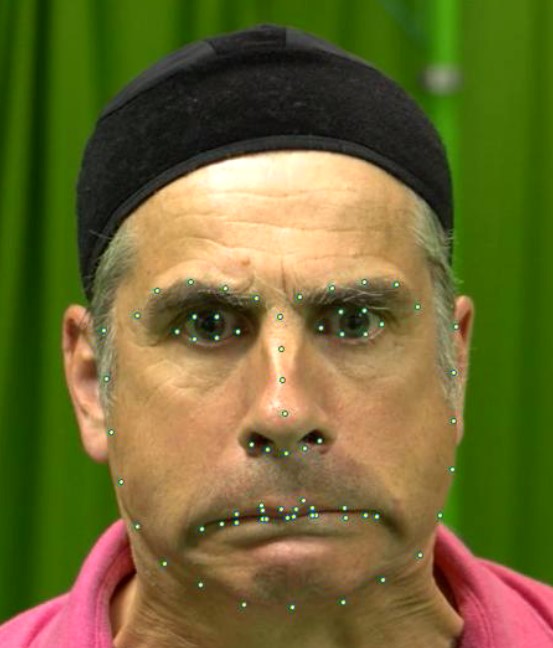}  
    	             
\caption{Landmark points initialized using an ensemble of regression trees and updated by the AAM}
	\label{fig:Landmarks}
\end{figure}

\begin{figure}[b]
    \centering        
    	   \includegraphics[width=\linewidth]{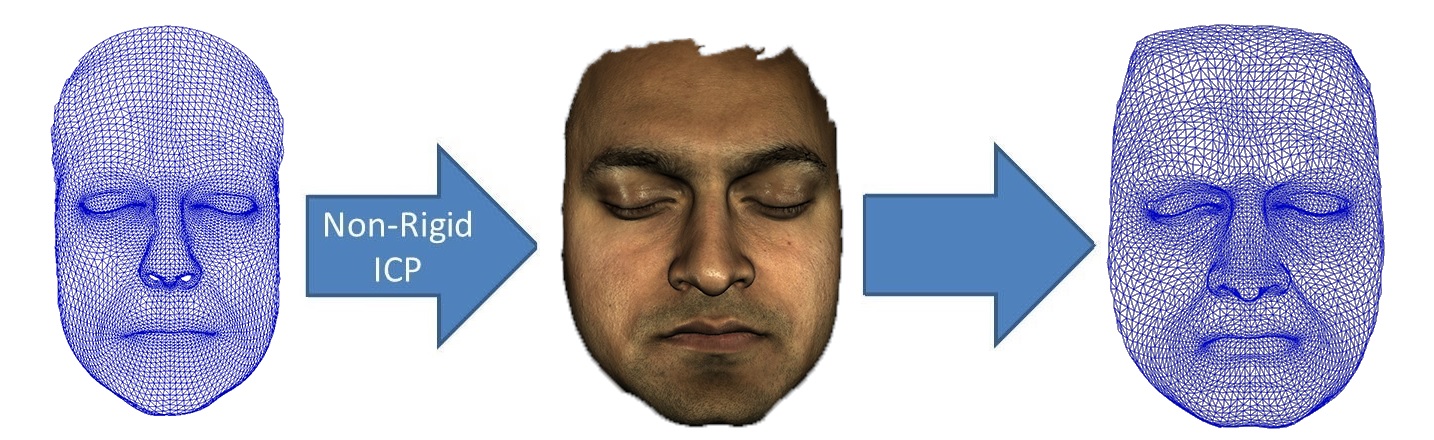}            
           \includegraphics[ height = 1in]{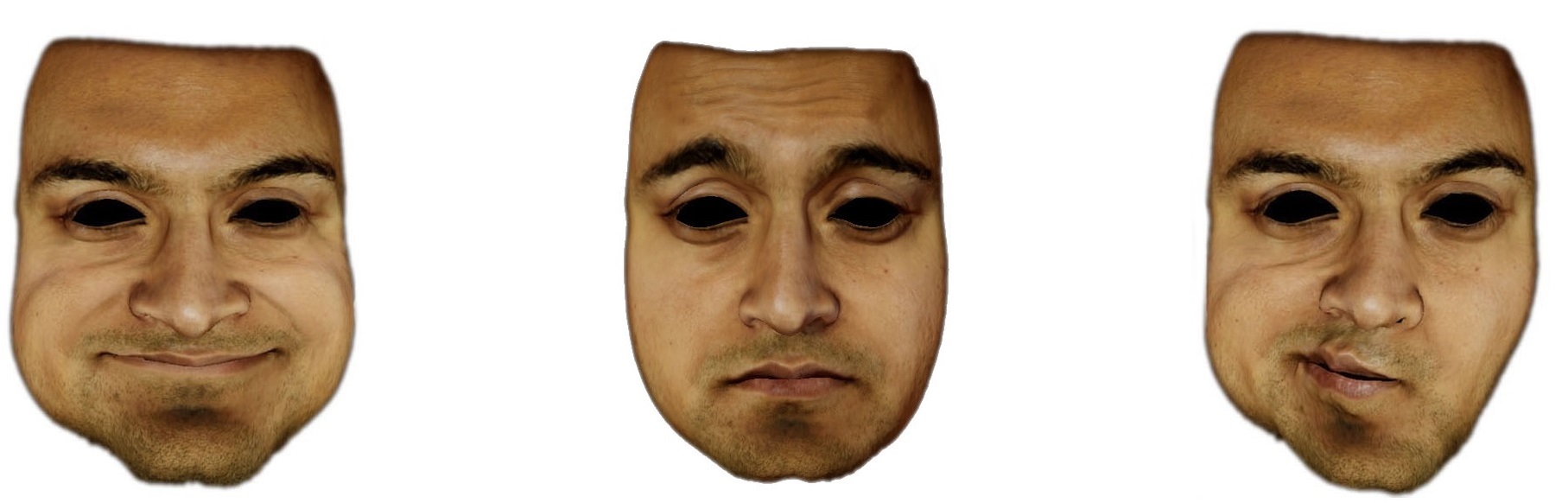}      	           
\caption{ Non-rigid ICP from template mesh to scan resulting in user specific Blendshapes}
	\label{fig:ICP}
\end{figure}
\section{Automatic Blendshape Generation}
\label{sec:AutomaticBlendshapeGeneration}
\noindent In order to obtain the neutral expression mesh of the actor, we scan the actor in a one-time pre-processing step. Alternatively, this can also be obtained automatically using the approach of \cite{BlanzVetter_1999_3DMorphableModels}. The resulting scanned mesh has a random noisy topology which cannot be used directly. In order to rectify this, we then deform the neutral expression mesh from an existing template Blendshape model to this scanned mesh using the algorithm of \cite{Amberg_2007_OptimalStepNonRigidICP} to obtain a new mesh with the desired topology. We then automatically generate the expression Blendshapes for the actor's face by applying the Deformation Transfer approach of \cite{SumnerPopovic_2004_DeformationTransfer} to give us 140 unique Blendshapes. This gives us a linear Blendshape model with $N$ Blendshapes for use in our solver as below:
\begin{eqnarray}
\label{eqn:Blendshapes}
{B} = {B}_0 + \sum_{i=1}^{N} \alpha_i ({B}_i - {B}_0)
\end{eqnarray}
\noindent where, $B_{0}$ is the neutral expression Blendshape, $\alpha_{i}$ is the weight associated with Blendshape $i$ and $B_{i}$ corresponds to the i-th Blendshape.

\noindent The vertices on the 3D mesh that correspond to the automatically detected landmarks from Sec.\ref{sec:2DFeatureDetection}, are chosen by the user in a one time manual step.

\section{Camera Calibration for Facial Projection}
\label{sec:CameraParameters}
\noindent We calibrate our camera using a standard checkerboard pattern in order to obtain our camera intrinsic matrix $K$ in a one time step. We then calculate the extrinsic parameters $[R|t]$ i.e. the rotation and translation that take the 3D face mesh from model coordinates to camera coordinates. This is essentially a Perspective-n-Point problem between the vertex coordinates on the mesh and the corresponding landmarks in 2D. Given these corresponding landmarks, we then solve for the camera extrinsic parameters every frame using the Levenberg-Marquadt non-linear optimization framework combined with RANSAC for robustness. This gives us the $[R|t]$ values every frame that we combine with the camera intrinsic matrix $K$ to give us our projection matrix -- $K[R|t]$. We update the values of the 3D landmark points on the mesh every frame by using the previous frame's shape. This ensures better projection every frame and improves the accuracy of the solver. In order to account for differences between adjacent frames, we further process these results and smooth them across frames by feeding these into the Kalman filter. 

\section{Online Facial Feature Smoothing}
\label{sec:KalmanFilter}

\noindent The 2D features and the camera extrinsic parameters so far, are obtained on a frame by frame basis. Both the 2D landmark detections and the camera extrinsic estimates are prone to noise and this independence between the frames inevitably leads to jitter in the final animation as shown in the accompanying video. This necessitates a smoothing operation in order to ensure consistency and avoid sudden changes.  We use a Kalman filter in order to smoothly transition these values between frames. The Kalman filter predicts the value of the landmarks and extrinsic parameters every frame and then uses the observations to update its belief about what the parameters should actually be, based on the value of the Kalman-gain-factor, which it calculates based on the noise in observation and the noise in the process. This ensures that any updates made to the landmarks and the extrinsic parameters are updated smoothly. We use one Kalman filter for updating the changes in all the 2D landmarks and one for updating the changes in rotation and translation parameters for the camera extrinsics.

\begin{figure*}[t]
    \centering
    	   \includegraphics[clip, trim=0 0.6in 0 0.8in,width=\textwidth]{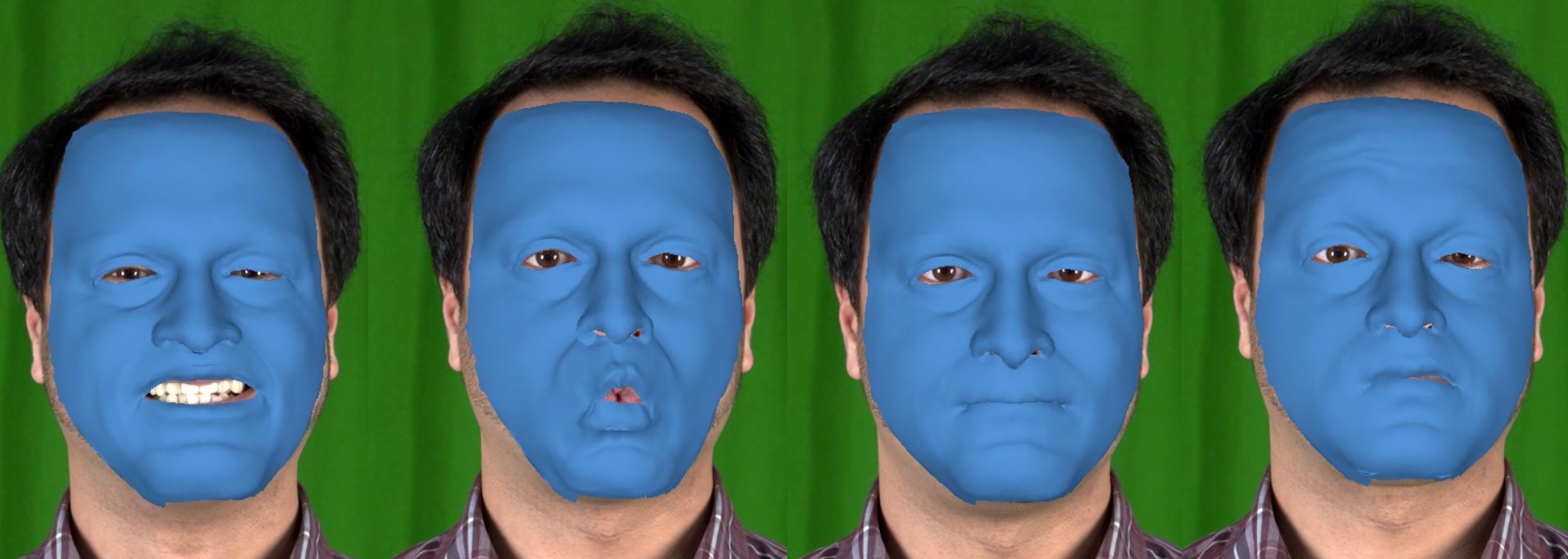}  
\caption{Results from the projection of our 3D face mesh onto the video}
	\label{fig:Projections}
\end{figure*}
\noindent The location of the landmark point detections are predicted every frame by taking into account the velocity and acceleration along each dimension of the tracked points based on the previous frames and then combined with the observations at that frame. A similar process is applied for the translation values in the camera extrinsics. As for the rotations, the orientations in 3D have 3 degrees of freedom ( yaw, pitch and roll ) but our rotation matrix has 9 parameters. Smoothing the rotation matrix directly in this space will not constrain the values properly and will not ensure valid rotation values. Hence we perform the smoothing in quaternion space. This removes ambiguity and also constrains the rotations better thus ensuring that our rotation values are valid throughout the sequence.

\noindent This process of online smoothing using Kalman filters greatly helps with tackling noise inherent in the 2D feature detection and camera extrinsics and ensures that our objective function isn't affected by noise. The results of our updated landmarks and camera projection matrix can be seen in Figure \ref{fig:Projections} and in the accompanying video.

\section{2D Objective Function}
\label{sec:2DObjectiveFunction}
\noindent Given the corrected 2D landmarks and the projection parameters from the previous step, we then solve for the facial expressions by calculating the optimal coefficient weights for the Blendshape model using the following objective function, similar to \cite{ChenCao_2013_3DShapeRegression}.
\begin{eqnarray}
\label{eqn:2DObjectiveFunction}
E_{2D(\alpha)} = \sum_{l=1}^{n} \space \|\Pi_{Q}(M(B_{0} + \sum_{i=1}^{N}\alpha_{i}B_{i})^{(v_{l})}) - q^{(l)} \|^{2}
\end{eqnarray}
\noindent where:
\begin{itemize}

\item $n$ is the number of landmarks
\item $\Pi_{Q}$ is the camera projection matrix 
\item $M$ is the rigid transform from object space to camera coordinates
\item $N$ is the number of Blendshapes
\item $B_{0}$ is the neutral expression Blendshape 
\item $\alpha_{i}$ is the weight associated with Blendshape $i$ with the constraint  $0 \leq \alpha_{i} \leq 1$
\item $B_{i}$ corresponds to the i-th Blendshape
\item $q^{(l)}$ represents the l-th 2D landmark point in the image
\item $v_{l}$ represents the vertex corresponding to landmark $l$

\end{itemize}

\noindent This objective function by itself only considers the 2D landmarks to reduce the error and is inherently under constrained as the information along the depth axis is missing in the video. This leads to insufficient constraining and to errors that show up especially along the depth axis. In Section \ref{sec:PriorConstraints} and \ref{sec:DataBinning} we tackle this problem by improving on our objective function and making use of prior constraints.

\section{3D Spatial Constraints}
\label{sec:PriorConstraints}

\noindent One of the problems inherent in monocular capture approaches is that the information along the depth dimension is lost. This leads to situations where the optimization function described above is able to obtain coefficients that minimize the squared distance between the projected vertices and the corresponding 2D landmarks but it does so with solutions that no longer adhere to physically plausible shapes of the face. This effect is especially visible around the mouth region as there is a lot of variation in the depth dimension during speech. This can also be seen when opening the jaw as there is movement along the depth axis. In order to ensure that our objective function provides physically plausible results, we need to ensure that our results are regularized to stay within such a solution space. In general 3D face capture systems can be less flexible compared to monocular markerless systems and placing physical markers on the actor can be very time-consuming, but these systems provide accurate tracking of points. There is ample data available from legacy 3D face capture systems from multiple people performing different facial expressions and speech sequences. It makes sense to use this data in order to regularize our results. We can use the data from these accurate 3D systems and use it to constrain our results while still retaining a monocular markerless approach. 

\noindent In a one time training step, we use prior data in the form of 3D marker locations over capture sequences from previous captures using the Vicon Cara 3D head-mounted device \cite{Vicon}. Our data spanned multiple sequences of speech and facial expressions from 5 different people performing diverse facial movements. In order to use this data for our purposes, we first need to solve for the Blendshape coefficient weights specific to our Blendshape model. We do this using the standard objective function \cite{HaoLi_2010_ExampleBasedFacialRigging} for 3D solves, as shown below.
\begin{eqnarray}
\label{eqn:3DObjectiveFunction}
E_{\mathrm{3D}(\alpha)} = \arg\min_{\alpha} \space  \| {B}_{0} + {B}\alpha  \ - {T} \|_{2}^{2} + \beta\|\alpha\|_{1}+ \alpha^{T}\Gamma\alpha 
\end{eqnarray}
where:
\begin{itemize}

\item $B_{0}$ is a $3n\times1$ vector representing the neutral face, where $n$ is the number of markers.

\item $B$ is a matrix of size $3n\times N$, that contains the deltas for each of the Blendshapes $B_{i...N}$, where $N$ is the number of Blendshapes.

\item $\alpha$ is a $N\times 1$ vector of weights with the constraint  $0 \leq \alpha_{i} \leq 1$.

\item $T$ is a $3n\times1$ vector representing the target markers.

\item The term $\|\alpha\|_{1}$ is an L1-norm on $\alpha$ that penalizes the sum of weights.

\item $\beta$ is a weighting factor on the L1 regularizer. 

\item The $\Gamma $ term is a Tikhonov regularizer that ensures that the function is convex and has a unique global solution. $\Gamma = \epsilon \, I$ where $\epsilon$ is a very small constant and $I$ is the identity matrix of size $N\times N$. 
\end{itemize}

\noindent Solving this for our multiple training sequences gives us valid coefficient weights over multiple people. We use this data $D_{prior}$, in order to estimate the posterior of the solution in a probabilistic framework. Essentially given a solution vector of coefficient weights i.e. our likelihood, we need to estimate the posterior $P(\alpha|D_{prior})$, given the prior $P(\alpha)$. We can learn this prior $P(\alpha)$ using a Kernel Density Estimation method - the Parzen window with an RBF kernel as explained below:
\begin{eqnarray}
\label{eqn:RBFKernel}
P(\alpha) = \frac{1}{n} \sum_{i=1}^{n} \frac{1}{\sqrt[]{2\pi\sigma}}\exp \left( - \frac{(\alpha_{i} - \alpha)^{2}}{2\sigma^{2}} \right)
\end{eqnarray}
where $n$ is the number of prior data points, $\alpha$ is the estimated coefficient vector and $\alpha_{i}$ are the prior points. The bandwidth or standard deviation $\sigma$ is obtained as shown in the next section.

\section{Data Binning}
\label{sec:DataBinning}

\noindent One of the issues with using a density estimation technique directly is that it will give higher probability to points that occur more frequently in our prior data $D_{prior}$. This is not desirable as we assume that all of our prior data is accurate as we obtained it from a high accuracy 3D capture system and we want to weight them equally. So in order to overcome this, we essentially need to perform a data-binning operation where we replace multiple data points by a single vector corresponding to the mean of the cluster to which they belong. This makes sense as it allows us to give equal weight to different valid facial configurations without letting its frequency affect the probability. We use the Mean Shift algorithm \cite{MeanShift_1975_Fukunaga} to cluster our prior data. 

\noindent The mean shift algorithm is a non-parametric clustering algorithm which does not require the knowledge about the number of clusters. It works by updating candidates for centroids to be the mean of the points within a given region. Given a candidate centroid $x_{i}$ for iteration $t$, the candidate is updated according to the following equation:
\begin{eqnarray}
\label{eqn:MeanShift}
x_{i}^{t+1} = x_{i}^{t} + m(x_{i}^{t})
\end{eqnarray}
\noindent where $m$ is the mean shift vector that is computed for each centroid that points towards a region of the maximum increase in the density of points

\noindent We make use of an automatic method of bandwidth selection \cite{AutomaticBandwidthSelection_Chacon_2012,FullBandwidthMatrix_Horova_2013,ComparisonBandwidthSelectors_Chacon_2013} for use with the mean shift algorithm. This gives us our bandwidth, $\sigma$ in equation \ref{eqn:RBFKernel}. We then use the means of the clusters thus obtained ($\alpha_{i}$) within the Parzen window density estimation framework (Eqn. \ref{eqn:RBFKernel}) in order to obtain the prior probability of a solution.

\section{Objective Function with 3D Spatial Prior}
\label{sec:ObjectiveFunctionWithConstraints}
\noindent Finally, we factor this probability into our objective function in order to obtain 
optimal coefficients $\alpha$ that adhere to physically plausible face configurations. Since the probability of the upper face coefficients should be independent of the lower face coefficients --- e.g. to avoid the probability of the eyebrows being raised being affected by the jaw being open --- we perform the density estimation of these coefficients separately. This is essential since our prior data is not guaranteed to cover all possible combinations of the upper-face and lower-face shapes in tandem. Our updated objective function is as shown below:
\begin{eqnarray}
\label{eqn:PriorObjectiveFunction}
E_{Final(\alpha)} = E_{2D} + \frac{\lambda}{P(\alpha|D_{prior})}
\end{eqnarray}
\noindent where $\lambda$ is the weight given to the probability term and $ P(\alpha|D_{prior})$ is the posterior distribution on the coefficients. This ensures that the solution provided by our objective function lies within the valid space of facial expressions which is controlled by the prior data. As seen in the results in Figures \ref{fig:Comparison2},\ref{fig:Comparison3} and \ref{fig:Comparison4}, this yields significant improvement over the solve using just the 2D landmarks. This is especially visible in regions around the mouth as this is where majority of the variation in depth occurs.

\section{Results and Discussion}


\begin{figure*}[h]
    \centering
           \includegraphics[width=\textwidth]{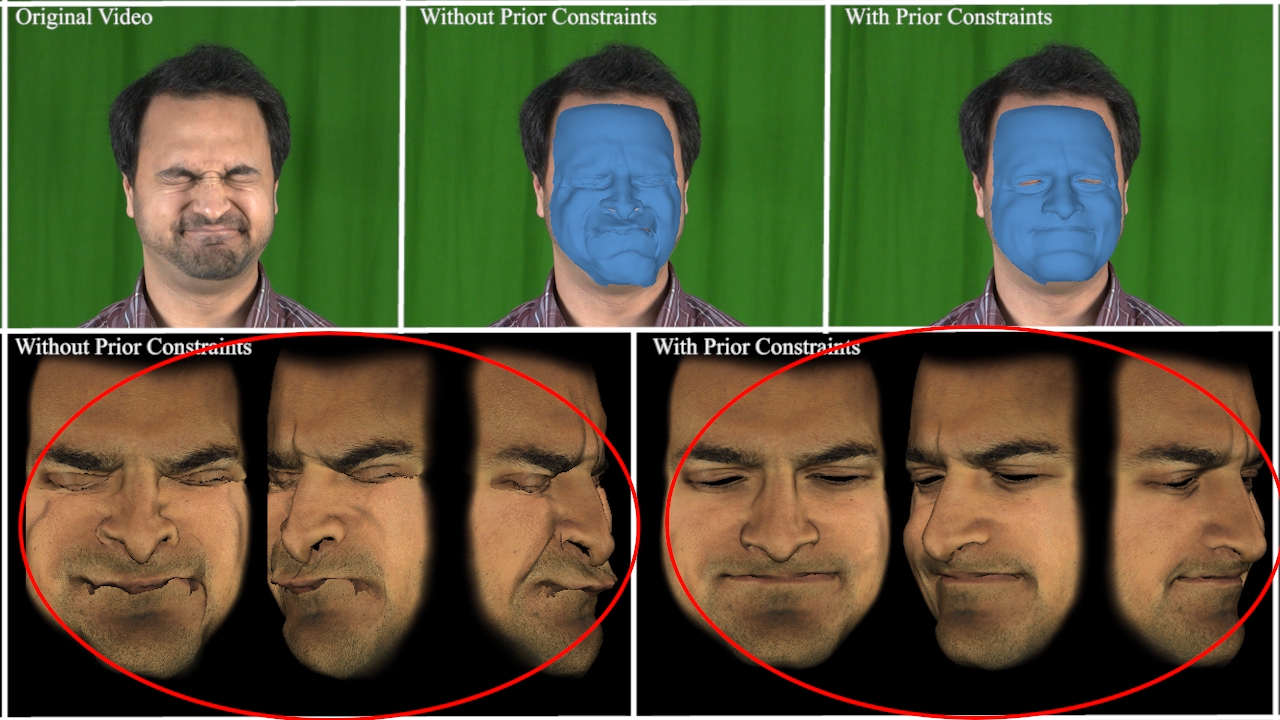} 
\caption{Result showing a comparison between the solve using only 2D points vs the improvements obtained using our Prior constraint.}
	\label{fig:Comparison2}
\end{figure*}

\begin{figure*}[h]
    \centering
    	   \includegraphics[width=\textwidth]{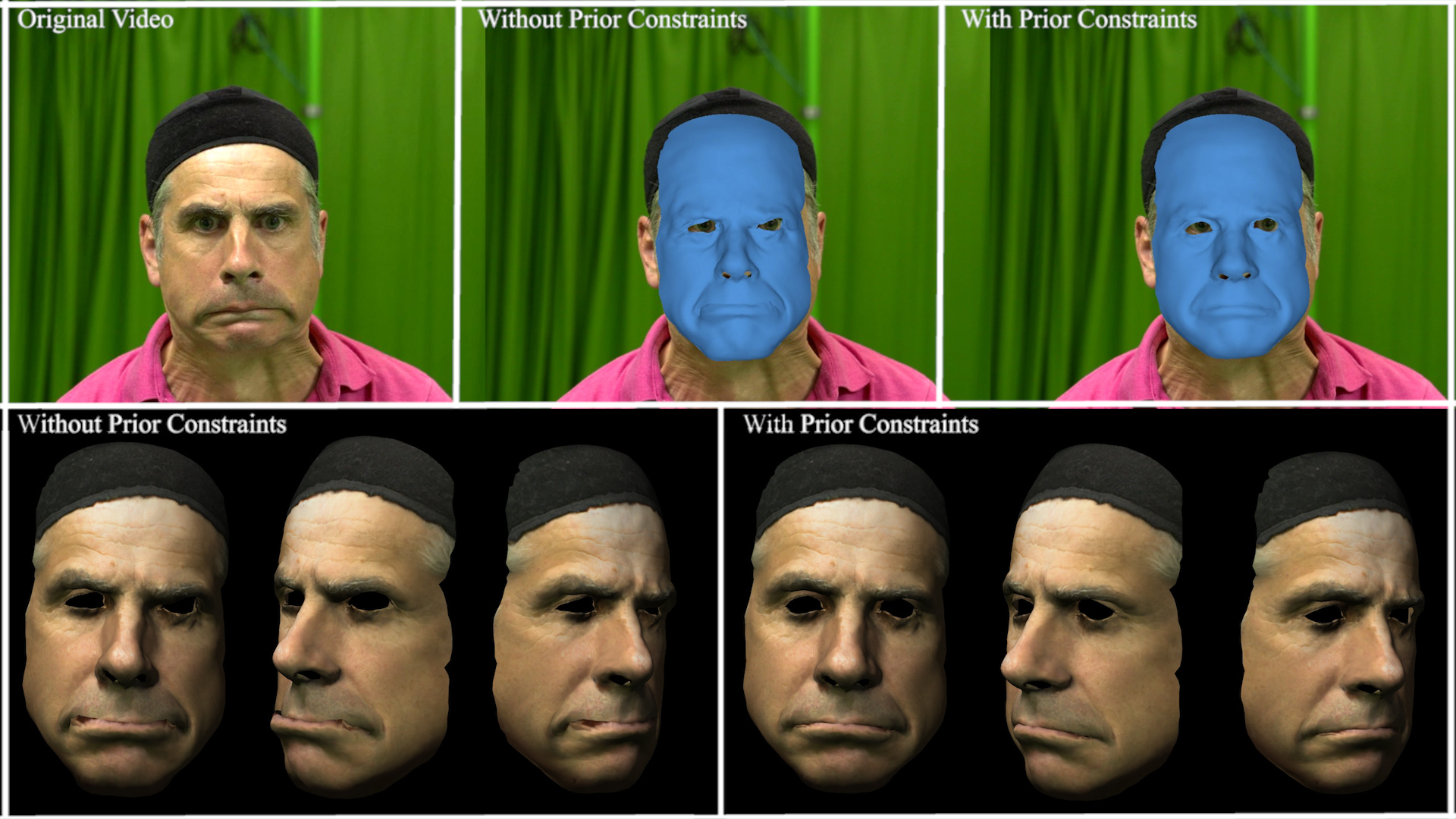}         
\caption{Comparison of the frown expression using only 2D points vs the improvements obtained using our Prior constraint.}
	\label{fig:Comparison3}
\end{figure*}

\begin{figure*}[h]
    \centering
    	   \includegraphics[width=\textwidth]{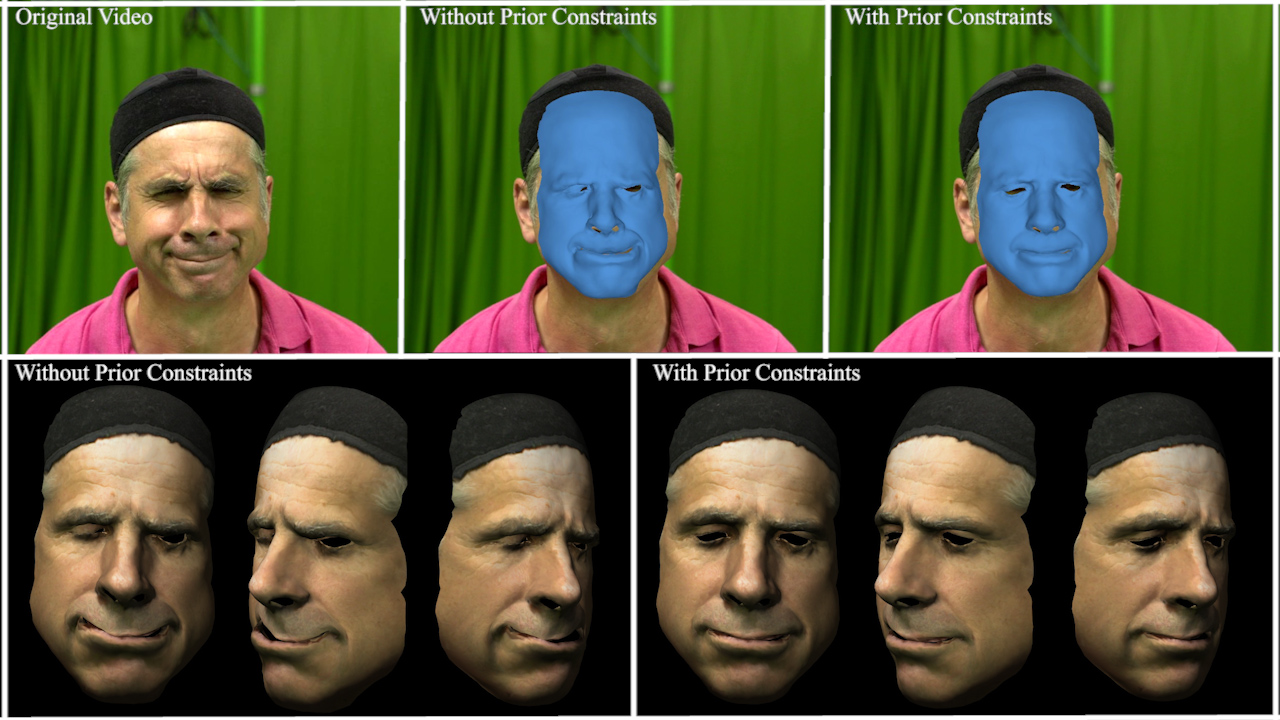}         
\caption{Comparison of the angry expression using only 2D points vs the improvements obtained using our Prior constraint.}
	\label{fig:Comparison4}
\end{figure*}


\begin{figure*}[h]
    \centering
    	   \includegraphics[width=\textwidth]{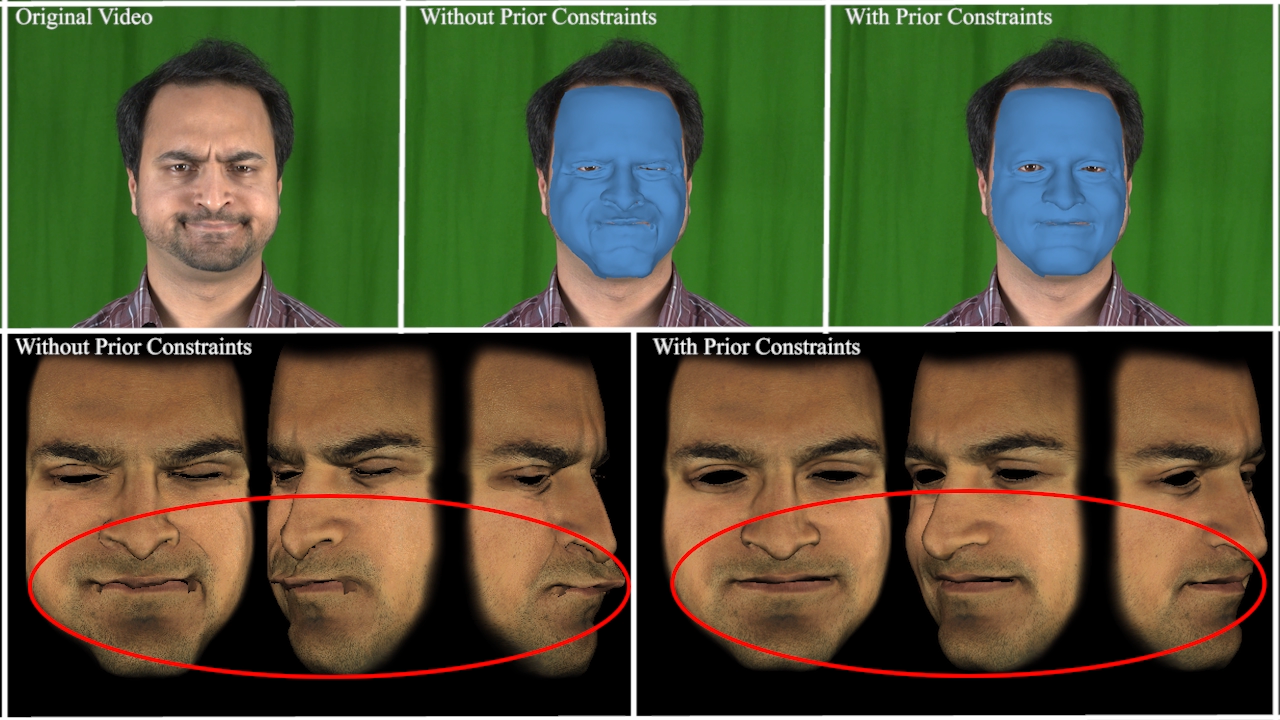}          
\caption{Result showing a comparison between the solve using only 2D points vs the improvements obtained using our Prior constraint.}
	\label{fig:Comparison5}
\end{figure*}

\noindent We render our 3D model within an OpenGL framework combined with OpenCV for calculation of the projection matrix and solving the perspective-n-point problem. In order to perform our optimization, we make use of the Ceres solver, an open source C++ library for modeling and solving constrained non-linear optimization problems. Our method is online as it doesn't require any operations across frames, although it isn't real-time owing to speed bottlenecks. Our system consists of a 4th generation Intel-I7 2.80GHz quadcore processor with 16GB RAM and an NVIDIA GTX765M graphics card.

While the 2D feature detector used for initializing the landmarks per frame is independent of the user (as it's trained on different faces), in our experiments it wasn't accurate enough, especially for extreme facial expressions including ones with occlusions. The Active Appearance Model is trained on 15 images of the user performing few facial expressions that elicit the range of his expressions both on the upper and lower face regions. These training images are marked semi-automatically -- initialized using the previous 2D feature tracker and then corrected where needed. Although this makes our system dependent on the specific actor, it is a small price to pay for the improvement in tracking especially considering only 15 images were sufficient to track accurately through a video sequence of over a thousand frames.

\noindent The accompanying video shows the improvements obtained owing to our Kalman filtering operation performed both on the 2D features and on the camera projection parameters.

\noindent As seen in the results in Figures \ref{fig:Comparison2}, \ref{fig:Comparison3}, \ref{fig:Comparison4} and \ref{fig:Comparison5}, the addition of the prior makes our results more reliable. The improvements can be seen in general but specifically around the mouth region where the changes in depth are most prominent. Areas where occlusions are common, around the eyes and mouth, stand to benefit the most as these regions are prone to loss of information in the 2D input. As our density estimator involves the Radial Basis Function as a kernel in the Parzen Window scheme, our objective function becomes non-linear, but because we initialize the parameters with the solutions from the previous frame, we do not run into popping in the animations owing to local minima. Nevertheless, the Parzen window density estimator is the prime bottleneck in taking this approach from merely online to real-time. In future work we will consider more efficient mechanisms for calculating prior probabilities such as Probabilistic PCA \cite{Tipping_PPCA_1999} or Gaussian Process Latent Variable models \cite{NeilLawrence_2004_GPLVM}. Our prior data consisted of animation sequences over 5 people and multiple facial expressions and speech movements to give us a total of 45817 frames of animation. Our automatic bandwidth estimation gives us a bandwidth of 0.3066 and 81 clusters for the lower face and a bandwidth of 0.1202 and 82 clusters for the upper face.

\subsection{Quantitative Evaluation - Comparison with Ground Truth}
\label{sec:Quant}

\begin{figure}[h]
    \centering          
    		\includegraphics[clip, trim=0.0in 0.0in 0.0in 0.0in,width=\linewidth]{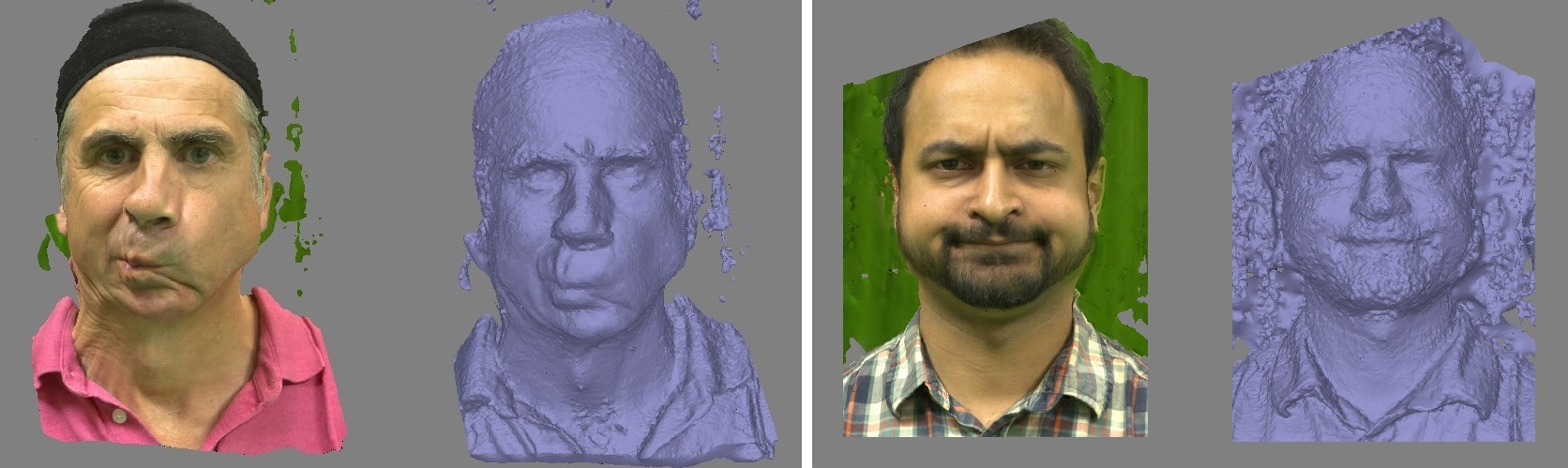}
\caption{Ground-truth geometry obtained by reconstructing synchronized frames from multiple views of the actor's face.}
	\label{fig:GroundTruthGeometry}
\end{figure}

\begin{figure}[!htb]
    \centering
    	   \includegraphics[clip, trim=1.2in 2.5in 1.2in 2.5in, width= \linewidth]{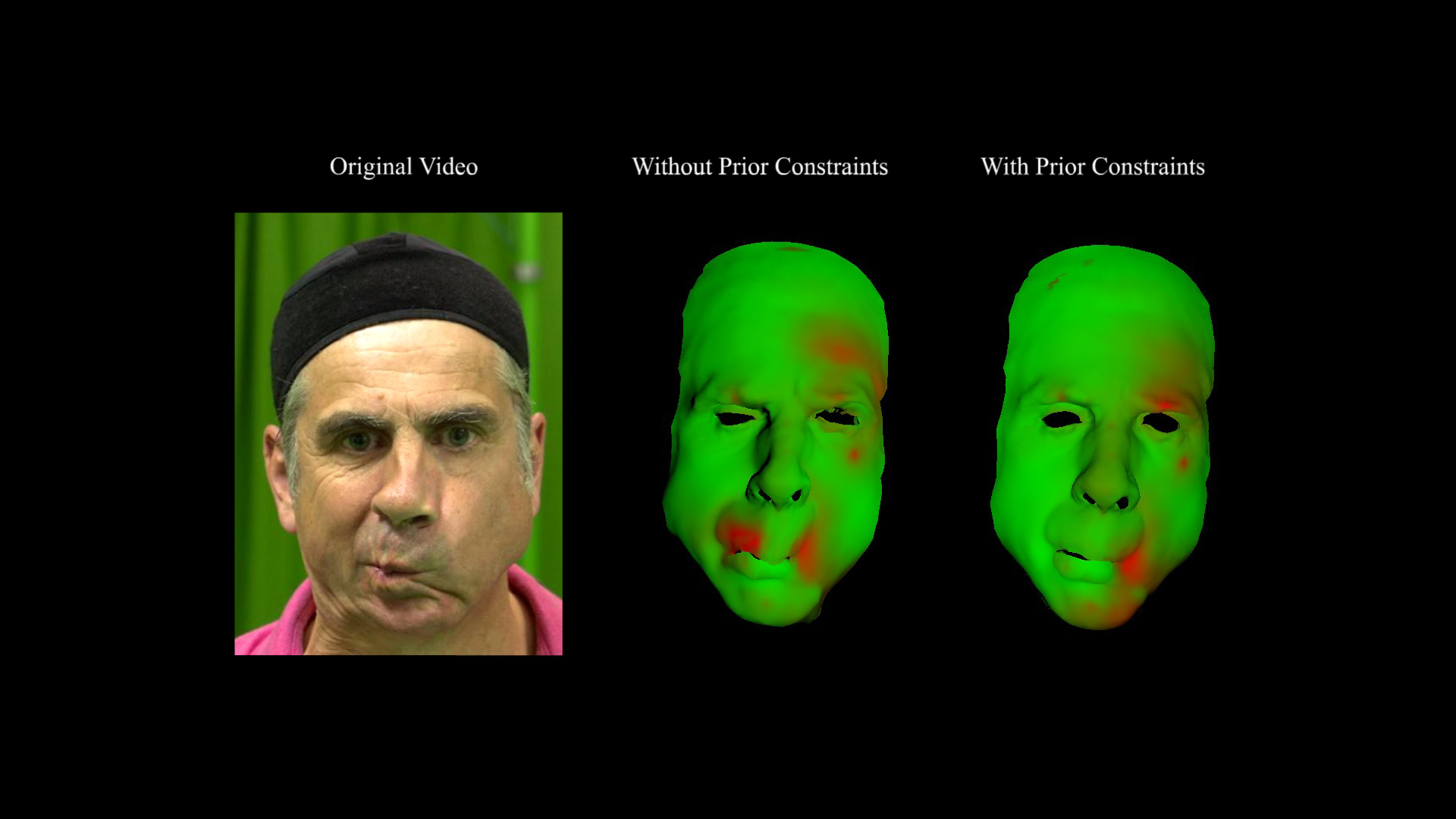}     
           \includegraphics[clip, trim=1.2in 2.5in 1.2in 2.5in, width= \linewidth]{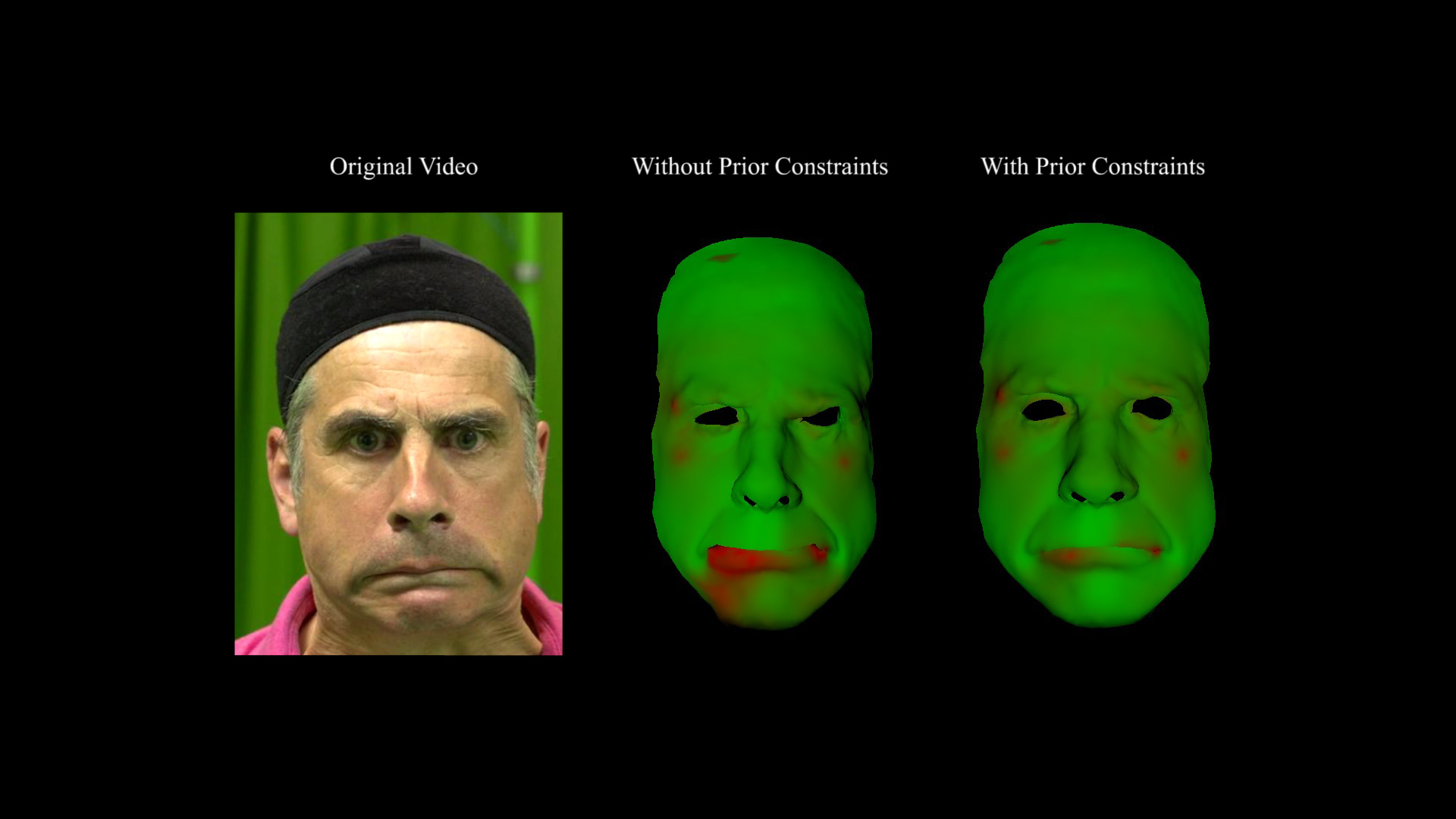} 
\caption{Ground-truth evaluation for the lip-swing expression (top) and the frown expression (bottom). The total error from ground truth geometry for the lip-swing expression was: 4.4908e+05 $cm^{2}$ (without prior) and 3.3752e+05 (with prior). The total error from ground truth geometry for the frown expression was: 5.5391e+05 $cm^{2}$ (without prior) and 4.3288e+05 (with prior). The ground truth geometry was reconstructed from 5 different views of the actor's face.}
	\label{fig:Heatmap1}
\end{figure}

\begin{figure}[!htb]
    \centering
    	   \includegraphics[clip, trim=1.2in 2.2in 1.2in 2.0in, width= \linewidth]{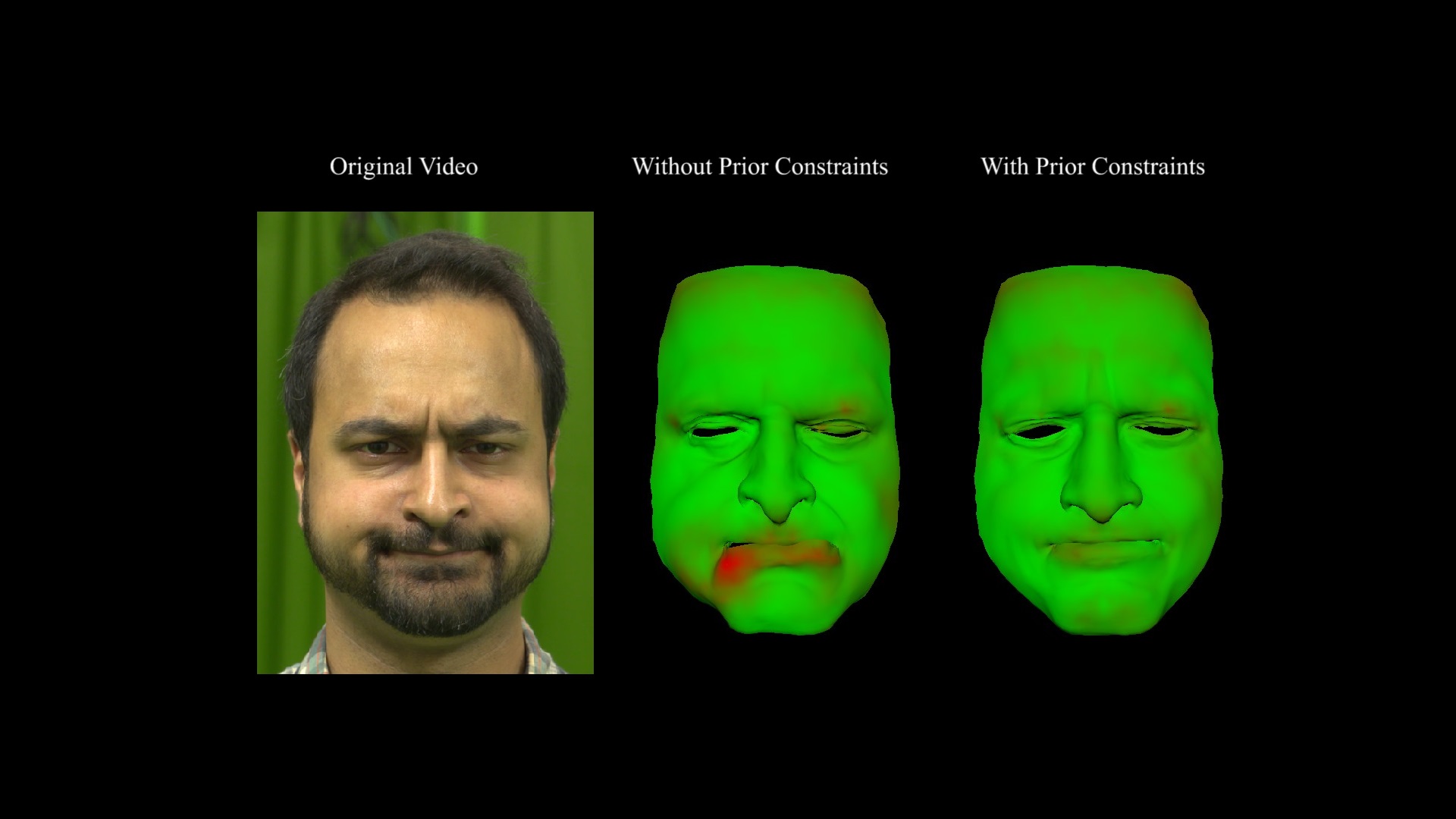}     
           \includegraphics[clip, trim=1.2in 2.2in 1.2in 2.0in, width= \linewidth]{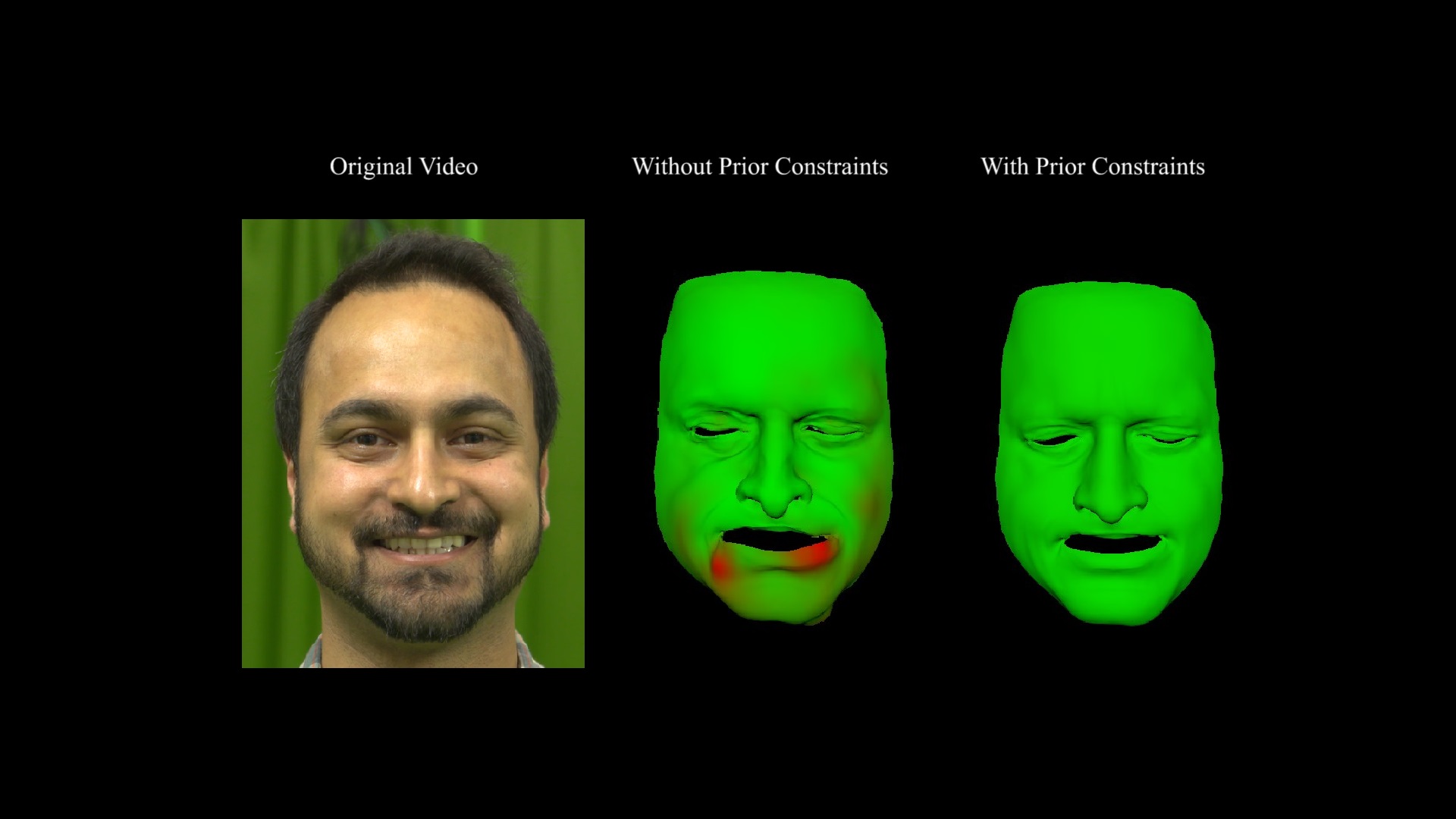} 
\caption{Ground-truth evaluation for the angry expression (top) and the smile expression (bottom). The total error from ground truth geometry for the angry expression was: 2.06719e+05 $cm^{2}$ (without prior) and 1.78974e+05 (with prior). The total error from ground truth geometry for the smile expression was: 8.18939e+05 $cm^{2}$ (without prior) and 3.47592e+05 (with prior). The ground truth geometry was reconstructed from 5 different views of the actor's face.}
	\label{fig:Heatmap2}
\end{figure}

In order to evaluate the results of our algorithm quantitatively, we needed to collect ground truth data for comparisons. For this purpose, we recorded both of our subjects using 5 high-definition cameras with baseline offsets that capture the actor's face from 5 different angles. These cameras were synced with each other using our gen-lock mechanism allowing us to obtain video frames from the performance sequence from different views that are temporally in correspondence with each other. We use the data from these cameras for a few expressions and reconstruct the 3D geometry of the actor's face for those frames using the Agisoft Photoscan software \cite{Agisoft_2014_agisoft} and obtain high detail textured meshes as shown in Figure \ref{fig:GroundTruthGeometry}. The obtained meshes have a random topology and hence we use the method of Amberg et al. \cite{Amberg_2007_OptimalStepNonRigidICP} to deform the neutral face of the actor to the reconstructed 3D meshes. This gives us ground truth geometry with the same topology as our Blendshapes which we use for quantitative evaluation. We then align the resulting shapes and the ground-truth geometry in 3D space by doing a rigid transformation after manually selecting corresponding points on the 3D meshes. We generate a heat map based on the per-vertex error of our results compared to the ground-truth. The results of this evaluation can be seen in Figure \ref{fig:Heatmap1} and \ref{fig:Heatmap2}. The heat map was generated by using the error value per vertex (after scaling between 0 and 1) as a UV coordinate on a color gradient image. The gradient image goes from green at 0 to red at 1. Regions that have higher error show up in red and regions with lower error show up in green. The resulting quantitative errors are as explained in the image captions in Figures \ref{fig:Heatmap1} and \ref{fig:Heatmap2}.

\section{Comparison with reduced Blendshape Set}
\label{sec:EmilyComparison}

In the previous sections, our Blendshape model was comprised of 140 Blendshapes. We note that one of the factors affecting the results of the solver is the fact that the Blendshapes themselves are not all compatible with each other i.e certain combinations of Blendshapes when triggered together result in a shape that is no longer within the feasible spectrum of shapes that a human face can assume. This is expected as the Blendshapes are not orthogonal to each other unlike in methods using PCA bases \cite{IanMatthews_2011_InteractiveRegionBased}. Our claim is that Blendshapes that are inherently more compatible with each other --- i.e. Blendshape sets containing bases such that there are fewer combinations of shapes that result in implausible expressions when triggered together --- will result in better solutions. Naturally, this implies that a Blendshape set having shapes that go relatively well with each other, will perform better. In order to test this hypothesis, we make use of the 'Emily' Blendshape set \cite{Alexander_2010_DigitalEmily}, which is comprised of 68 shapes, and solve for the Blendshape weights using our solver without the prior constraints and then quantitatively evaluate it with the results obtained using our original 140 Blendshapes, with and without the prior constraints.

\begin{figure}[!htb]
    \centering
    	   \includegraphics[clip, trim=0.0in 0.0in 0.0in 0.0in, width= \linewidth]{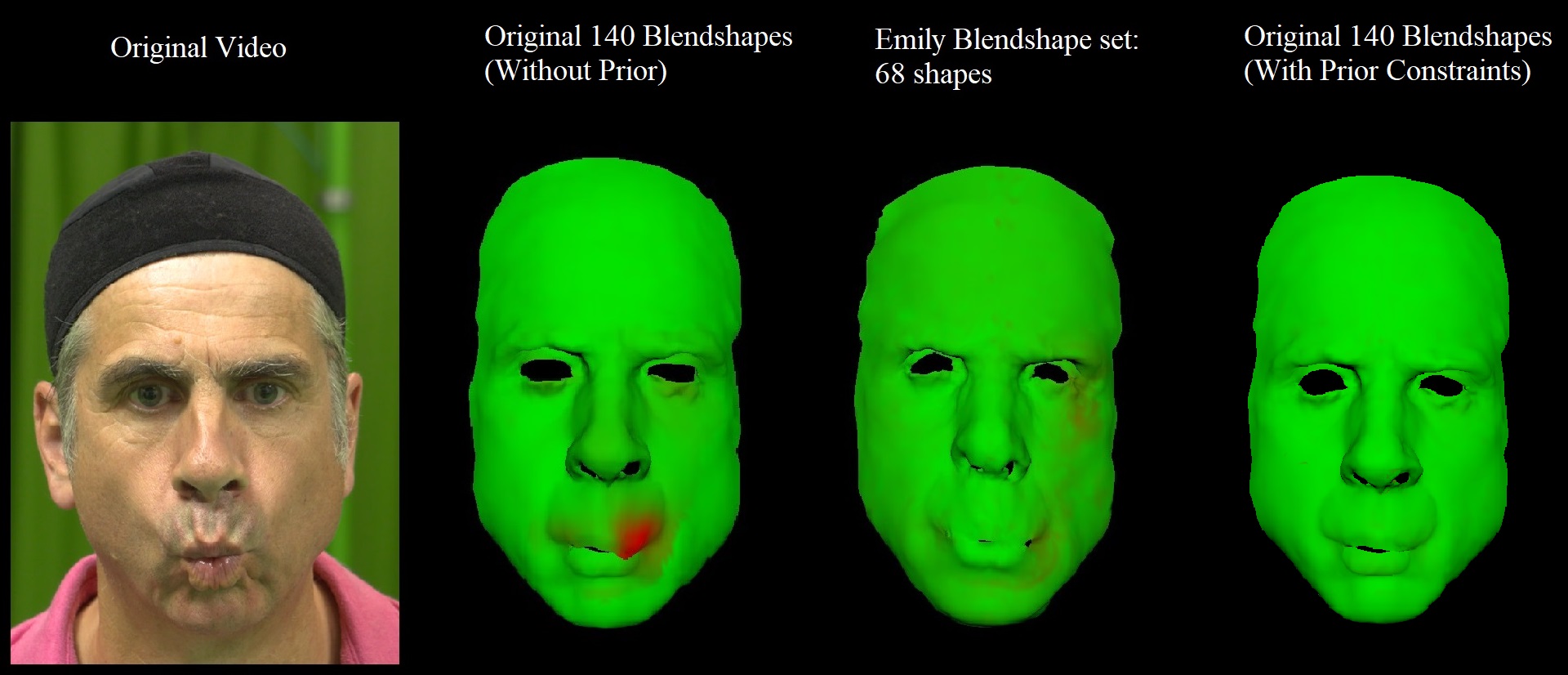}    
           \includegraphics[clip, trim=0.0in 0.0in 0.0in 0.0in, width= \linewidth]{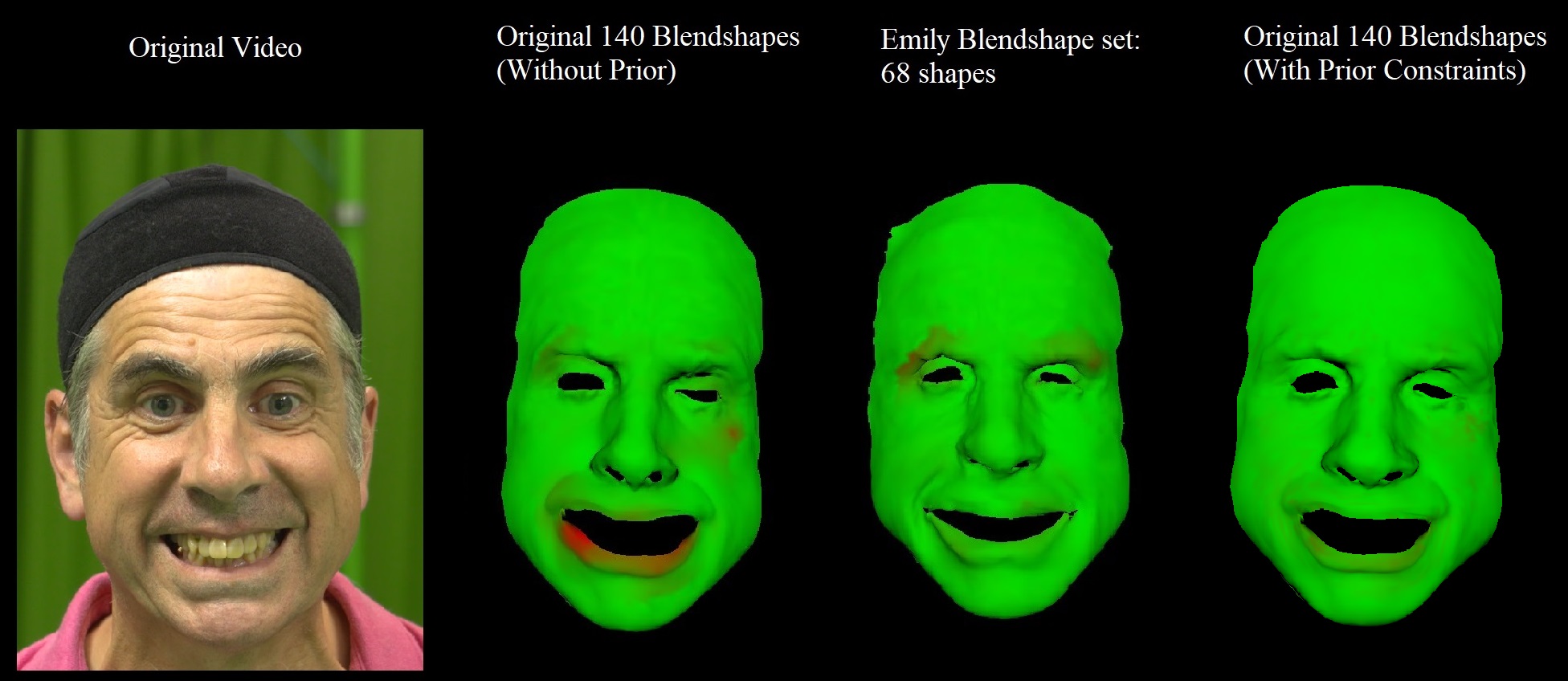} 
\caption{Ground-truth evaluation for the OO expression (top) and the smile expression (bottom). The total error from ground truth geometry for the OO expression was: 1.53622e+05 $cm^{2}$ (140 Blendshapes, without prior), 4.6528+04 $cm^{2}$ (140 Blendshapes, with prior constraints) and 1.37559e+05 $cm^{2}$ (Emily Blendshapes).The total error from ground truth geometry for the smile expression was: 1.36572e+05 $cm^{2}$ (140 Blendshapes, without prior), 6.6960e+04 $cm^{2}$ (140 Blendshapes with, prior constraints) and 1.04310e+05 (Emily Blendshapes).}
	\label{fig:Heatmap3}
\end{figure}

Note that because the two Blendshape sets are inherently different from each other, we restrict the error calculation to the relevant facial regions comprising of the frontal face, excluding the outer vertices around the edges of the mesh. This is to ignore the error resulting from the edge vertices which occurs due to the different sizes of and the number of faces in the two Blendshape sets. Once again, we ensure that the topology of both the resulting Emily shapes and those from our original Blendshape set are identical by using the method of \cite{Amberg_2007_OptimalStepNonRigidICP} and follow it by a per-vertex error calculation. The results of these comparisons can be seen in Figure \ref{fig:Heatmap3}. As can be seen, the error using the Emily set of Blendshapes is lower than those using our original 140 Blendshapes when we solve for both without using the prior constraints, but the results when we solve with the prior constraints are significantly lower in error when compared to the previous results. This shows that while the choice of Blendshape sets makes a difference, incorporating our prior in the solver provides significantly better solutions.

\section{Conclusion}

\noindent We have presented a lightweight markerless approach for capturing the facial movements of a user using only a 2D monocular input resulting in high quality animation parameters. We make use of prior constraints on the solve results, obtained from pre-existing accurate 3D captures and use this to improve the physical plausibility of our solve results which helps tackle the loss of information in depth that is inherent in 2D monocular inputs. We combine 2D landmark detections based on an ensemble of regression trees with an Active Appearance Model for improved accuracy. We make use of Kalman filters to handle noise across frames in an online fashion, both in 2D feature detection and in estimating the camera parameters giving us improved tracking and solves. We quantitatively compare the results of our solver which uses the spatial constraints with one that doesn't and show that our method generates better results when compared to the ground truth data. We also evaluate the effect that the choice of Blendshapes has on the results and quantitatively show that while the choice of shapes has an impact, overall our solution using the spatial prior still generates results closer to the ground truth data.
\clearpage

\bibliographystyle{eg-alpha-doi}
\bibliography{Bibliography}

\end{document}